\documentclass{sn-jnl}

\usepackage{manyfoot}%
\usepackage{fullpage}
\usepackage{algorithm}%
\usepackage{algorithmicx}%
\usepackage{algpseudocode}%

\floatname{algorithm}{Algorithm}

\usepackage{graphicx}
\usepackage{amsmath}
\usepackage{psfrag}
\usepackage[sort&compress,comma,square,numbers]{natbib}
\usepackage{color,amssymb}
\usepackage{amsthm}
\theoremstyle{plain}%
\newtheorem{theorem}{Theorem}
\newtheorem{corollary}{Corollary}
\newtheorem{definition}{Definition}%
\usepackage{thmtools, thm-restate}

\providecommand{\sct}[1]{{\texttt{#1}}}
\newcommand{\Dcor}{\sct{Dcor}_{n}}
\newcommand{\MDcor}{\sct{MDcor}_{n}}
\newcommand{\ADcor}{\sct{ADcor}_{n}}
\newcommand{\Dcov}{\sct{Dcov}_{n}}

\newcommand{\Xn}{\mathbf{X}}
\newcommand{\Yn}{\mathbf{Y}}
\newcommand{\mbx}{\ensuremath{X}}
\newcommand{\mby}{\ensuremath{Y}}
\providecommand{\mc}[1]{\mathcal{#1}}
\newcommand{\EE}{\mathbb{E}}
\newcommand{\trace}[1]{{\ensuremath{\operatorname{trace}\!\left(#1\right)}}} 

\begin{document}

\title[Article Title]{High-Dimensional Independence Testing via Maximum and Average Distance Correlations}

\author[1]{Cencheng Shen}
\author[2]{Yuexiao Dong}
\affil[1]{Department of Applied Economics and Statistics, University of Delaware}
\affil[2]{Department of Statistics, Operations, and Data Science, Temple University}

\abstract{This paper investigates the utilization of maximum and average distance correlations for multivariate independence testing. We characterize their consistency properties in high-dimensional settings with respect to the number of marginally dependent dimensions, compare the advantages of each test statistic, examine their respective null distributions, and present a fast chi-square-based testing procedure. The resulting tests are non-parametric and applicable to both Euclidean distance and the Gaussian kernel as the underlying metric. To better understand the practical use cases of the proposed tests, we evaluate the empirical performance of the maximum distance correlation, average distance correlation, and the original distance correlation across various multivariate dependence scenarios, as well as conduct a real data experiment to test the presence of various cancer types and peptide levels in human plasma.}

\keywords{unbiased distance correlation, chi-square test, testing independence}
\maketitle

\section{Introduction}
\label{sec:intro}
Given pairs of observations $(x_{i},y_{i}) \in \mathbb{R}^{p} \times \mathbb{R}^{q}$ for $i=1,\ldots,n$, assume they are independently identically distributed as $F_{\mbx \mby}$ of finite moments. The statistical hypothesis for testing independence is formulated as:
\begin{align*}
&H_{0} : F_{XY} = F_{X}F_{Y}, \\
&H_{A} : F_{XY} \neq F_{X}F_{Y}.
\end{align*}
Traditional correlation measures like Pearson's correlation \citep{Pearson1895} are commonly used but unable to detect nonlinear and high-dimensional dependencies. Recent measures, such as the distance correlation \citep{SzekelyRizzoBakirov2007,SzekelyRizzo2009} and the Hilbert-Schmidt independence criterion \citep{GrettonEtAl2005,GrettonGyorfi2010}, can uncover any type of dependency given sufficient sample size, are asymptotically zero if and only if independence, and can consistently test independence for any joint distribution of fixed dimensionality. Due to such consistency properties, dependence measures have become widely used in various applications, such as feature screening \citep{LiZhongZhu2012,Zhong2015,DCorScreening,Zhang2025}, time-series \citep{Zhou2012,Pitsillou2018,DCorTemporal}, conditional independence \citep{Gretton2007,SzekelyRizzo2014,Wang2015}, K-sample test \citep{Edelmann2022,DCorMANOVA}, clustering \citep{Szekely2005,Rizzo2010}, graph testing \citep{MGCGraph,DCorGraph}, and deep learning \citep{Guo2022, Zhen2022}.

Detecting multivariate dependencies, and especially in high-dimensional scenarios, remains a challenging task with limited understanding. As the number of dimensions increases relative to the sample size, the testing power of existing dependence measures may diminish \citep{RamdasEtAl2015}. In situations where the dimensions ($p$ or $q$) approach infinity and the sample size grows more slowly than the dimension, distance correlation may fail to detect certain multivariate dependencies \citep{Shao2019}. To address this issue, a common solution involves projecting the high-dimensional data into a low-dimensional space and then considering the marginal dependence, such as the average marginal covariance \citep{Shao2019}, projection correlation \citep{Zhu2017}, and average covariance of random rotations \citep{Huo2017}.


In this paper, we examine the utilization of maximum distance correlation and average distance correlation as sample test statistics for multivariate dependence testing. We formulate the maximum distance correlation and the average distance correlation based on pairwise, unbiased, and marginal distance correlations. To understand their respective advantages, we establish their consistency properties for high-dimensional independence testing, using the concept of marginally dependent dimensions. Subsequently, we analyze their limiting null distribution, and propose a valid chi-square-based test for computing p-values, which is significantly faster than the standard permutation test. Our numerical study compares the performance of maximum, average, and original distance correlation using both Euclidean distance and the Gaussian kernel across various simulation settings. Finally, we provide a real data experiment on cancer types and peptide levels in human plasma to illustrate their practical applications. All theorem proofs and additional simulations are in the appendix.

\section{Background}

In this section, we provide a review of existing results, including the unbiased distance correlation, its relationship with the Hilbert-Schmidt independence criterion (HSIC), its validity and consistency in testing independence, the standard permutation test, the limiting null distribution, and the chi-square test. We denote the paired sample data as follows:
\begin{align*}
(\mathbf{X},\mathbf{Y}) = \{(x_{i},y_{i}) \in \mathbb{R}^{p+q}, i=1,\ldots,n\}.
\end{align*}
Namely, $(\mathbf{X},\mathbf{Y})$ represents sample realizations of the random variable $(X,Y)$, where each sample pair $(x_{i},y_{i})$ is assumed to be independently and identically distributed as $F_{XY}$. Throughout this paper, we assume finite moments for $F_{XY}$.

\subsection*{Distance Correlation}
Given a distance metric $d(\cdot,\cdot)$ such as the Euclidean metric, we denote $\mathbf{D}^{\mathbf{X}}$ as the $n \times n$ pairwise distance matrix of $\mathbf{X}$ with $\mathbf{D}^{\mathbf{X}}_{ij}=d(x_i,x_j)$. Similarly, we denote $\mathbf{D}^{\mathbf{Y}}$ as the pairwise distance matrix of $\mathbf{Y}$. 
Next, we compute a modified matrix $\mathbf{C}^{\mathbf{X}}$ as follows:
\begin{align*}
\mathbf{C}^{\mathbf{X}}_{ij}=
 \begin{cases}
 &\mathbf{D}^{\mathbf{X}}_{ij}-\frac{1}{n-2}\sum\limits_{t=1}^{n} \mathbf{D}^{\mathbf{X}}_{it}-\frac{1}{n-2}\sum\limits_{s=1}^{n} \mathbf{D}^{\mathbf{X}}_{sj} \\
 &\ \ \ \ \ +\frac{1}{(n-1)(n-2)}\sum\limits_{s,t=1}^{n}\mathbf{D}^{\mathbf{X}}_{st}, \ i \neq j \\
 &0, \mbox{ otherwise},
 \end{cases}
\end{align*}
and similarly compute $\mathbf{C}^{\mathbf{Y}}$ from $\mathbf{D}^{\mathbf{Y}}$. The unbiased sample distance covariance and correlation are then given by:
\begin{align*}
& \Dcov(\mathbf{X}, \mathbf{Y}) = \frac{1}{n(n-3)}\trace{\mathbf{C}^{\mathbf{X}}\mathbf{C}^{\mathbf{Y}}},\\
&\Dcor(\mathbf{X},\mathbf{Y})= \frac{\Dcov(\mathbf{X},\mathbf{Y})}{\sqrt{\Dcov(\mathbf{X},\mathbf{X})\Dcov(\mathbf{Y},\mathbf{Y})}} \in [-1,1].
\end{align*}
If $n<4$ or the denominator term is not a positive real number, the unbiased sample distance correlation is set to $0$. 

\subsection*{Theoretical Property}

The above unbiased statistic was introduced in \cite{SzekelyRizzo2014}. Comparing to the biased statistic introduced in \cite{SzekelyRizzoBakirov2007}, the unbiased statistic satisfies
\begin{align*}
\EE(\Dcor(\mathbf{X},\mathbf{Y})) = 0
\end{align*}
if and only if $X$ and $Y$ are independent. 

By default, distance correlation utilizes the Euclidean distance as its metric. Nevertheless, it may accommodate any distance metric or kernel choice by setting $\mathbf{D}^{\mathbf{X}}$ and $\mathbf{D}^{\mathbf{Y}}$ as the corresponding distance or kernel matrices. It is worth noting that when the Gaussian kernel is used, distance correlation effectively becomes equivalent to HSIC. In fact, one can interchange between distance and kernel metrics through an appropriate kernel-to-distance transformation \citep{SejdinovicEtAl2013,DCorKernel}.

When the metric used is of strong negative type \citep{Lyons2013,Lyons2018}, such as the Euclidean distance, or when a characteristic kernel is used \citep{GrettonEtAl2005,Gretton2007}, like the Gaussian kernel, the resulting distance correlation exhibits the following property:  
\begin{align*}
\Dcor(\mathbf{X},\mathbf{Y}) \stackrel{n \rightarrow \infty}{\rightarrow} 0
\end{align*}
if and only if $X$ and $Y$ are independent. 
This fundamental property makes distance correlation a valid and universally consistent statistic for testing independence. 

\subsection*{Hypothesis Test}
To conduct testing on sample data, one needs to approximate the null distribution of the test statistic, typically achieved through a standard permutation test \citep{GoodPermutationBook}. Given $\mathbf{X}$ and $\mathbf{Y}$, the process involves randomly permuting the indices of $\mathbf{Y}$ for $R$ iterations. The p-value is then computed as:
\begin{align*}
\mbox{pval}= \frac{1}{R}\sum_{r=1}^{R} 1\{\Dcor(\mathbf{X},\pi_r(\mathbf{Y}))> \Dcor(\mathbf{X},\mathbf{Y})\}
\end{align*}
where each $\pi_r$ represents a random permutation of size $n$. Specifically, when $X$ and $Y$ are dependent, performing a permutation test on distance correlation yields an asymptotic p-value of $0$, leading to testing power that converges to $1$ as the sample size $n$ increases. Conversely, when $X$ and $Y$ are independent, the p-value follows a \mbox{Uniform} distribution in the range $[0,1]$, and the testing power equals the type I error level $\alpha$. The primary drawback of the permutation test is its computational complexity, which requires repeated computation of the statistic for $R$ iterations, often set to $100$ or more, rendering it time-consuming for large sample sizes.

Recent works have significantly reduced the computational cost of distance correlation \citep{Huo2016,Hu2018} and the resulting permutation test \citep{DCorFast,Zhang2025}. Specifically, the null distribution of distance correlation satisfies the following \citep{zhang2018}:
\begin{align*}
n \cdot \Dcor(\mathbf{X},\mathbf{Y}) &\stackrel{D}{\rightarrow} \sum\limits_{i,j=1}^{\infty} w_{ij} (\mc{N}_{ij}^{2}-1),
\end{align*}
where the weights satisfy $w_{ij} \in [0,1]$ and $\sum_{i,j=1}^{\infty} w_{ij}^{2} = 1$, and $\mc{N}_{ij}$ are independent standard normal distributions. The null distribution can be further bounded by a chi-square-based distribution, enabling the use of the following chi-square test: 
\begin{align*}
\mbox{pval}=1-F_{\chi^{2}_{1}-1}(n \cdot \Dcor(\mathbf{X},\mathbf{Y})).
\end{align*}
This approximation is approximately valid for any $\alpha<0.0875$ \citep{DCorFast}, and has a time complexity of $O(1)$ and is straightforward to implement in any programming language. 

\subsection*{Difference from Mutual Independence}

A related but fundamentally distinct concept is that of testing mutual independence within a random vector \( X \). For instance, a recent paper \citep{Shao2019} introduced a similar idea of using the maximum of marginal statistics to construct a more powerful test for mutual independence. However, testing independence between \( X \in \mathbb{R}^p \) and \( Y \in \mathbb{R}^q \) is conceptually and technically different from testing mutual independence among the marginals of \( X \). While there are similarities in the construction of test statistics (e.g., via marginal aggregations), methods developed for one problem are generally not applicable to the other.

To be precise, testing mutual independence within \( X \) corresponds to the null hypothesis:
\[
H_0: F_X = \prod_{j=1}^p F_{X^j},
\]
i.e., all coordinates of \( X \) are mutually independent. This is a much stronger requirement than pairwise independence, and implies that there is no interaction or shared randomness among any subset of the components of \( X \).

By contrast, testing independence between \( X \) and \( Y \) involves the hypothesis:
\[
H_0: F_{XY} = F_X \cdot F_Y,
\]
which concerns the relationship between two random vectors, regardless of any dependence structure within \( X \) or within \( Y \) individually.

To illustrate the distinction, consider the combined vector \( Z = [X, Y] \). If \( X \perp Y \), this does not imply that the components of \( X \) are mutually independent — hence, \( Z \) may or may not be mutually independent. Conversely, if \( X \) and \( Y \) are dependent, then \( Z \) must be mutually dependent. As a result, applying a mutual independence test to \( Z \) is not a valid or consistent method for testing whether \( X \perp Y \), because mutual dependence of \( Z \) does not uniquely determine the presence or absence of dependence between \( X \) and \( Y \). The reverse also holds: tests for independence between \( X \) and \( Y \) are not valid nor consistent for assessing mutual independence within \( X \). 

The only special case in which the two problems coincide is when both \( X \) and \( Y \) are univariate. In that setting, testing independence between \( X \) and \( Y \) is equivalent to testing mutual independence within \( Z = [X, Y] \), which forms the basis for using marginal dependence measures for mutual independence testing in the bivariate case, as in \citep{Shao2019}. Outside of this setting, however, the two problems remain fundamentally different.

Therefore, while both settings involve notions of statistical dependence, they target distinct structural properties of the data, and the associated testing methodologies are not interchangeable in general.

\section{Main Results}

\subsection{Maximum and Average Distance Correlations}
Given $\Xn=\{x_i \in \mathbb{R}^{p} \mbox{ for }  i =1,\ldots, n]\}$ as the sample data, let $\Xn^s=\{x_i^s  \in \mathbb{R} \mbox{ for }  i =1,\ldots, n\}$ denote the $s$th-dimension of the sample data. Similarly for $\Yn$ and $\Yn^t$ for each $t \in [q]$. For every $s \in [p]$ and $t \in [q]$, we refer to the distance correlation between $(\Xn^s, \Yn^t)$ as the marginal distance correlation, denoted by $\Dcor(\Xn^s,\Yn^t)$. To distinguish, we will refer to $\Dcor(\Xn,\Yn)$, which incorporates all dimensions of the sample data, as the original distance correlation. 

We introduce the maximum and average distance correlations as follows:
\begin{align*}
&\MDcor(\Xn,\Yn)=\max\limits_{s \in [p],t\in[q]}\Dcor(\Xn^s,\Yn^t),\\
&\ADcor(\Xn,\Yn)=\sum\limits_{s \in [p],t\in[q]}\Dcor(\Xn^s,\Yn^t)/pq.
\end{align*}
In essence, $\MDcor(\Xn,\Yn)$ is the maximum of all marginal distance correlations per dimension, while $\ADcor(\Xn,\Yn)$ is the average of all marginal distance correlations. Note that in this paper, all the sample statistics utilize the unbiased distance correlation.

\subsection{Consistency for Testing Marginal Dependence}

In this section, our objective is to determine when and how the proposed statistics are suitable for testing independence in high-dimensional scenarios, particularly when $pq$ is large and increases concurrently with $n$. To achieve this, we introduce the concept of marginal dependence:\\

\begin{definition}
For $X \in \mathbb{R}^{p}$ and $Y\in \mathbb{R}^{q}$, we define $\Delta(X,Y)$ as the set of pairwise marginally dependent dimensions. In other words, the element $(s,t) \in \Delta(X,Y)$ if and only if $F_{X^s Y^t} \neq F_{X^s}F_{Y^t}$. We use $\delta(X,Y)=|\Delta(X,Y)|$ to denote the cardinality of this set, which means it represents the number of marginally dependent dimensions.\\
\end{definition}

Clearly, $\delta(X,Y) \in [0, pq]$. If $\delta(X,Y)>0$, it implies that $X$ and $Y$ must be dependent. However, the reverse is not always true; that is, $X$ and $Y$ may be dependent, while $\delta(X,Y)$ could be $0$. One such example is provided in appendix~\ref{appenSim}. In practice, creating such examples requires special construction, and the concept of marginal dependence captures a significant portion of dependence. 

The following two theorems establish the high-dimensional behavior of the maximum and average statistics under the null and alternative hypotheses, respectively.\\

\begin{restatable}{theorem}{thmOne}
\label{thm1}
Given $\delta(X,Y)= 0$, the average distance correlation satisfies:
\begin{align*}
\ADcor(\Xn,\Yn) &\stackrel{n \rightarrow \infty}{\rightarrow} 0 
\end{align*}
regardless of $pq$. On the other hand, the maximum distance correlation satisfies:
\begin{align*}
\MDcor(\Xn,\Yn) &\stackrel{n \rightarrow \infty}{\rightarrow} 0
\end{align*}
when $pq=o(\sqrt{n} e^{n})$.\\
\end{restatable}

\begin{restatable}{theorem}{thmTwo}
\label{thm2}
When $\delta(X,Y)> 0$, the maximum distance correlation satisfies:
\begin{align*}
\MDcor(\Xn,\Yn) \stackrel{n \rightarrow \infty}{\rightarrow} c >0
\end{align*}
regardless of $pq$. On the other hand, the average distance correlation satisfies:
\begin{align*}
\ADcor(\Xn,\Yn) \stackrel{n \rightarrow \infty}{\rightarrow} c \geq 0,
\end{align*}
with equality holds when $\delta(X,Y)=o(pq)$ and $pq \stackrel{n \rightarrow \infty}{\rightarrow} \infty$.\\
\end{restatable}

Therefore, both the maximum and average distance correlations are asymptotically consistent for testing the presence of marginal dependence when $pq$ is fixed because either statistic tends to zero asymptotically if and only if $\delta(X,Y)= 0$. However, in high-dimensional testing scenarios where $pq$ increases concurrently with $n$, the maximum statistic may not be consistent if $pq$ increases too rapidly relative to $n$, while the average statistic is not consistent when $\delta(X,Y)$ is too small relative to $n$. This can be summarized in the following corollary, which directly follows from Theorem~\ref{thm1} and Theorem~\ref{thm2}.\\

\begin{corollary}
As $n$ increases to infinity, the average distance correlation is asymptotically consistent in testing the existence of marginal dependence when $pq$ is fixed or $\delta(X,Y) =O(pq)$ for increasing $pq$.

The maximum distance correlation is asymptotically consistent in testing the existence of marginal dependence when $pq$ is fixed or $pq=o(\sqrt{n} e^{n})$ for increasing $pq$.\\
\end{corollary}

Therefore, the maximum correlation is advantageous when the dependence signal is sparse, i.e., concentrated in a few marginal pairs. In contrast, the average statistic is more powerful when there are many weak associations spread across many dimensions. 

\subsection{Limiting Null Distribution}
While one could employ the permutation test on either the maximum or average statistic (detailed procedure in appendix), it tends to be very slow for large $n$. To expedite the testing process, we consider the null distributions of the maximum and average statistics:\\

\begin{restatable}{theorem}{thmThree}
\label{thm3}
Assume that \( \mbx \) and \( \mby \) are independent, and that the coordinates of \( \mbx \) and \( \mby \) are independent within each variable. For sufficiently large $n$ and sufficiently small $\alpha$, it holds that
\begin{align*}
n \cdot \MDcor(\Xn,\Yn) &\preceq_{\alpha}  U-1,
\end{align*}
where $U \sim F_{\chi_{1}^{2}}^{pq}(x)$.\\
\end{restatable}

While the null distribution of maximum statistic relies on upper-tail dominance, the null distribution of the average statistic is simpler, which converges to a normal distribution as the dimensions increase:\\

\begin{restatable}{theorem}{thmFour}
\label{thm4}
Assume that \( \mbx \) and \( \mby \) are independent, and that the coordinates of \( \mbx \) and \( \mby \) are independent within each variable. As both $n$ and $pq$ increase to infinity, it holds that
\begin{align*}
n \sqrt{pq} \cdot \ADcor(\Xn,\Yn) \stackrel{D}{\rightarrow} N(0,2).
\end{align*}
\end{restatable}

Note that this coincides with the limiting null distribution of the original distance correlation in high-dimensions \citep{SzekelyRizzo2013a}. Moreover, while Theorem~\ref{thm3} holds for any $pq$, Theorem~\ref{thm4} requires $pq$ to increase. In practice, we have found that $pq >30$ suffices for a good approximation. Alternatively, if we assume each marginal distance correlation actually follows $\chi_{1}^{2}$, then the null distribution of $n pq \cdot \ADcor(\Xn,\Yn)$ equals $\chi_{pq}^{2}-pq$. This provides a better empirical approximation for small $pq$, and as $pq$ increases, it also converges to $N(0,2)$ after dividing by $\sqrt{pq}$.

Regarding the assumptions, we require that the coordinates of \( \mbx \) and \( \mby \) are independent. While this may appear restrictive at first glance, it can be further relaxed. Specifically, if we additionally assume that both \( p \) and \( q \) tend to infinity, the independence assumption can be relaxed to exchangeability, similar to Theorem 1 and Corollary 2 in \citep{SzekelyRizzo2013a}. This can be justified by de Finetti’s theorem, which states that an infinite exchangeable sequence is conditionally i.i.d. Our proofs remain valid under this setting, as the maximum or average distance correlation retains the same asymptotic distribution when conditioning on the latent variable.

\subsection{Chi-square-based Tests}
Utilizing the null distributions, we can construct chi-square-based tests for both the maximum and average distance correlations. Specifically, we calculate p-values as follows:
\begin{itemize}
\item For the maximum distance correlation, we let $z=n \cdot \MDcor(\Xn,\Yn)+1$, and compute the p-value as
\begin{align*}
\mbox{pval} = 1-Prob\{\chi^{2}_{1}<z\}^{pq}.
\end{align*}
If $\mbox{pval}<\alpha$, where $\alpha$ is the type 1 error, we reject the independence hypothesis.
\item For the average distance correlation, we let $z=n \cdot \ADcor(\Xn,\Yn) +1$, and compute the p-value as
\begin{align*}
\mbox{pval}=1-Prob\{\chi_{pq}^{2}<pq z\}.
\end{align*}
If $\mbox{pval}<\alpha$, where $\alpha$ is the type 1 error, we reject the independence hypothesis.
\end{itemize}
Note that for the average distance correlation, we employ the $\chi_{pq}^{2}$ distribution instead of the normal distribution. This choice provides a better approximation for small values of $pq$ while being equivalent to the normal distribution for large $pq$.

Both of these chi-square-based tests are considered valid according to the following theorem:\\

\begin{restatable}{theorem}{thmFive}
\label{thm5}
Under the same condition in Theorem~\ref{thm3}, the chi-square test for the maximum correlation is a valid test of independence for sufficiently large $n$ and sufficiently small type 1 error level $\alpha$. 
Moreover, the chi-square test for the average correlation is a valid test of independence for sufficiently large $n$ and $pq$, at any type 1 error level $\alpha$.\\
\end{restatable}

It is important to note that our results do not rely on any particular choice of distance metric. For instance, one can employ the Gaussian kernel and compute the maximum and average HSIC, and the chi-square-based tests remain approximately valid and consistent. 

\subsection{Computational Complexity}

Taken together, the maximum and average distance correlation statistics have a computational complexity of \( O(pq \, n \log n) \). This is because each marginal distance correlation can be computed in \( \mathcal{O}(n \log n) \) time under the Euclidean distance metric, thanks to \citep{Huo2016,Hu2018}. The chi-square-based tests incur no additional cost, so the overall complexity for testing remains \( O(pq \, n \log n) \). However, if a permutation test with \( R \) permutations is used, the total complexity increases to \( O(R \, pq \, n \log n) \), and in practice, \( R \) is typically 100 or more.

If one instead uses a non-Euclidean metric such as the Gaussian kernel, fast univariate algorithms are no longer available. In this case, the computation of the test statistic has a complexity of \( O(pq \, n^2) \), and using a permutation test raises the total complexity to \( O(R \, pq \, n^2) \).

For comparison, the original (non-marginal) distance correlation or kernel-based methods require \( O((p + q) \, n^2) \) time. However, since the pairwise distance or kernel matrix is computed only twice, the cost is typically treated as \( O(n^2) \) in practice. Therefore, marginal-based methods can be slower in terms of raw computation. Namely, in high-dimensional testing scenarios, potential gain in statistical power shall come at the cost of increased computational burden.

As an additional note, our average distance correlation performs similarly to the average distance covariance proposed in \citep{Shao2019}, since distance correlation is simply a normalized version of distance covariance, and their testing power has been shown to coincide \citep{SzekelyRizzoBakirov2007}. However, our method offers a computational advantage: the average distance covariance typically requires a permutation test (costing \( O(R \, pq \, n \log n) \)), or alternatively, a transformation into a \( t \)-distribution for avoiding permutations. The latter approach requires computing marginal distance variances, resulting in a complexity of \( O((pq + p^2 + q^2) \, n \log n) \), which can exceed the cost of a permutation test when $p>>q$ or vice versa.

In contrast, our average distance correlation, paired with the efficient chi-square test, achieves similar testing power and significantly lower computational cost, making it more suitable for large sample sizes and high-dimensional settings.



\section{Simulation Study}

In our simulation study, we first demonstrate that the chi-square-based distribution provides an accurate approximation of the true null distribution. Then we evaluate the testing power of the maximum and average tests on a variety of multivariate dependence.

\subsection{Chi-Square vs True Null}
\label{sim1}
\begin{figure*}[htbp]
\centering
\includegraphics[width=1.0\textwidth,trim={1cm 0cm 1cm 0cm},clip]{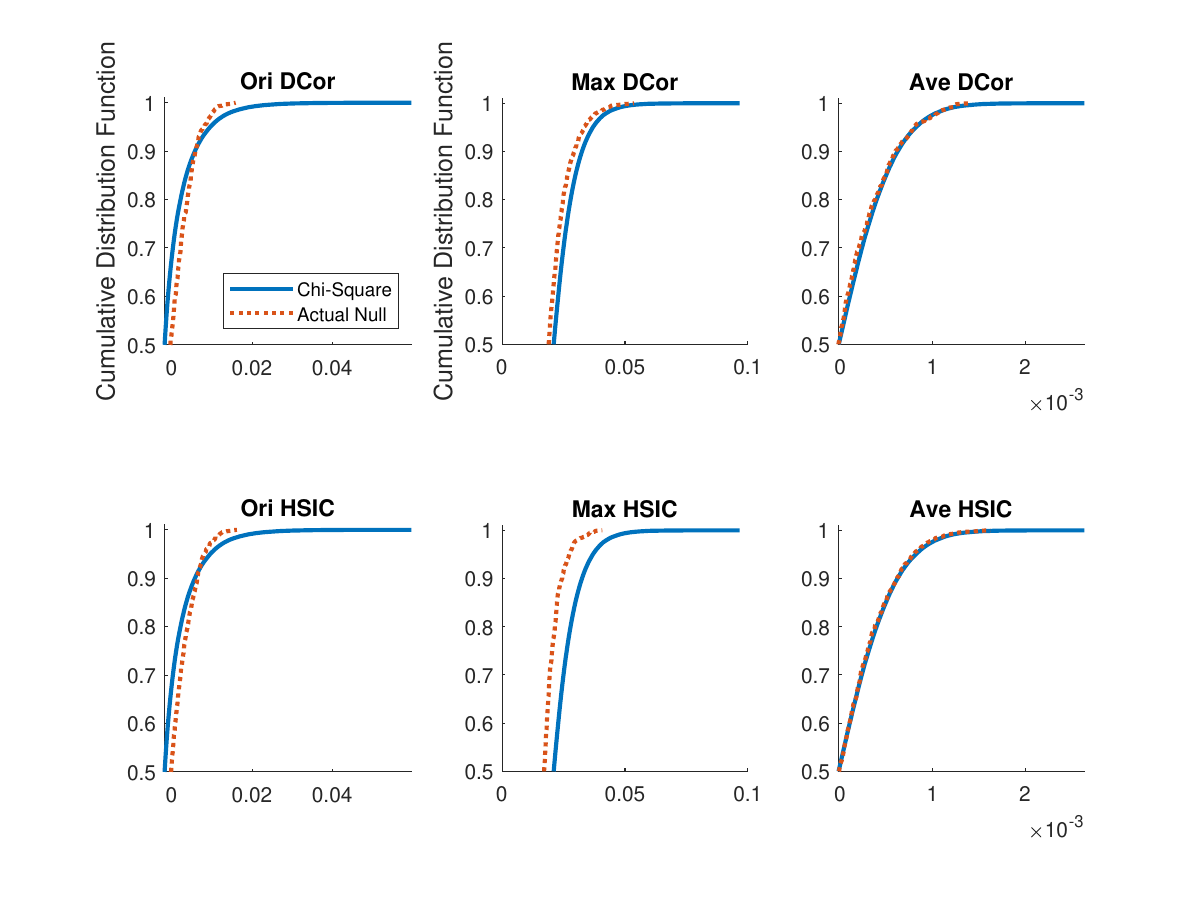}
\caption{Compare the chi-square distribution and true null distribution for distance correlation. The top row considers the Euclidean distance (DCor), while the bottom row employs the Gaussian kernel (HSIC). In the first column, we compare the null approximation for the original statistic. In the second and third columns, we compare the null approximation for the maximum and average statistics, respectively, based on Theorems~\ref{thm3} and~\ref{thm4}. }
\label{fig1}
\end{figure*}

Figure~\ref{fig1} displays the comparison between the chi-square-based distribution and the true null distribution for original, maximum, and average statistics, using Euclidean distance and Gaussian kernel respectively. The cumulative distribution function is plotted based on Theorem~\ref{thm3} and Theorem~\ref{thm4} for a sample size of $n=300$ and $pq=100$. The true null distribution is obtained through repeated generation of independent $(\Xn, \Yn)$. For the original and maximum distance correlations, the chi-square-based distribution dominates the true null for small $\alpha$, regardless of whether Euclidean distance or Gaussian kernel is used. For the average distance correlation, the chi-square-based distribution aligns closely with the true null distribution.

\subsection{Fixed $\delta(X,Y)$ with Increasing $pq$}
We evaluate the testing power of original statistic, maximum statistic, and average statistic, in detecting multivariate dependence structures. The data is generated by sampling $\mbx^s \sim \mbox{Uniform}(-1,1)$ for $s=1,\ldots,p$, using the $p \times 1$ vector $w=[1,\frac{1}{2},\frac{1}{3},\frac{1}{4},\frac{1}{5},0,\cdots,0]$, and considering
\begin{itemize}
\item Linear $(\mbx,\mby)$: $\mby =\mbx \cdot w$. 
\item Quadratic $(\mbx,\mby)$: $\mby=\mbx^2 \cdot w$ using dimension-wise square.
\item Fourth Root $(\mbx,\mby)$: $\mby=|\mbx|^\frac{1}{4} \cdot w$.
\item Independence $(\mbx,\mby)$: $\mbx^s \sim \mbox{Uniform}(-1,1)$ for $s=1,\ldots,p$ and $\mby \sim \mbox{Uniform}(-1,1)$.
\end{itemize}
Here, each sample in \( \Xn \) is independently and identically distributed (i.i.d.) from the distribution of \( X \), and \( \Yn \) is constructed as a function of each corresponding sample in \( \Xn \), except in the case of independence, where each sample in \( \Yn \) is also independently generated.

\begin{figure*}[htbp]
\centering
\includegraphics[width=1.0\textwidth,trim={1cm 0cm 1cm 0cm},clip]{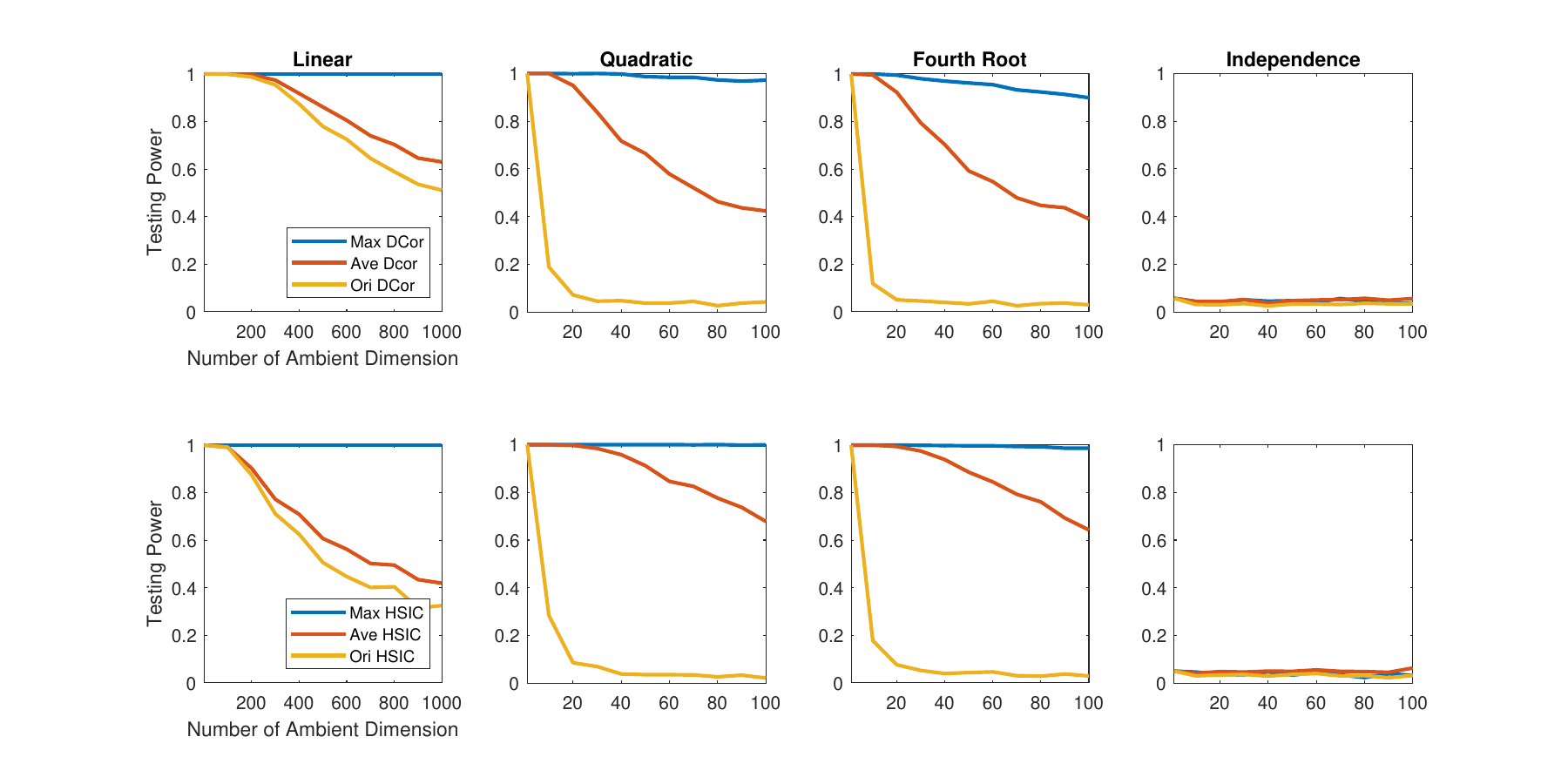}
\caption{Compare the testing power of maximum statistic, average statistic, and original statistic in linear, quadratic, fourth root, and independent settings as the number of dimensions increases while the number of marginally dependent dimensions is fixed. The top row utilizes Euclidean distance (DCor), while the bottom row employs the Gaussian kernel (HSIC).}
\label{fig2}
\end{figure*}

We perform the simulation study with $n=100$ by gradually increasing $p$ from $5$ to $100$ (except in the linear case, where it is increased to $1000$). At each $p$, we generate sample data $1000$ times, run each method and record the number of times the p-value is below $\alpha=0.05$. The results are plotted in Figure~\ref{fig2}, showing the testing power for each method.

In these scenarios (excluding independence), the number of marginally dependent dimensions are limited, i.e., $\delta(X,Y)=5$, while $pq$ increases. The maximum test delivers near perfect power, followed by the average test, and the original test has the lowest power. As $pq$ increases, the power of all tests declines, but the maximum test appears to be the least affected by increasing dimensions. This pattern remains consistent regardless of whether the Euclidean distance or Gaussian kernel is employed.

Furthermore, in the case of independence, the chi-square-based test effectively controls the type 1 error, and the test power closely aligns with $\alpha$, affirming the validity of the tests.

\subsection{Increasing $\delta(X,Y)$ with Fixed $pq$}

\begin{figure}[htbp]
\centering
\includegraphics[width=0.7\textwidth,trim={0cm 0cm 0cm 0cm},clip]{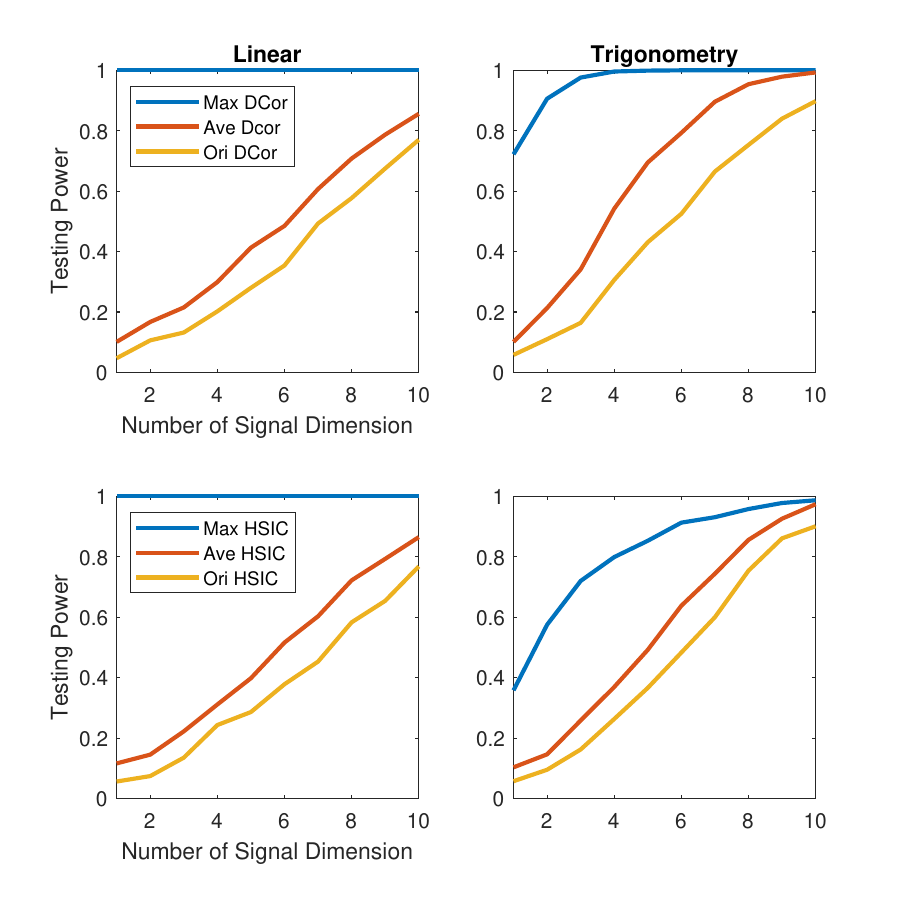}
\caption{Compare the testing power of maximum statistic, average statistic, and original statistic in linear and trigonometric relationships as the number of marginally dependent dimensions increases from $1$ to $10$. The top row considers the Euclidean distance (DCor), while the bottom row employs the Gaussian kernel (HSIC).} 
\label{fig3}
\end{figure}

In this scenario, we examine different multivariate dependence structures where the number of marginally dependent dimension $\delta(X,Y)$ increases, while $pq$ remains fixed. Let $\mbx^s \sim \mbox{Uniform}(-1,1)$ for $s=1,\ldots,p$, and consider
\begin{itemize}
\item Linear $(\mbx,\mby)$: $\mby^s =\mbx^s$ for each $s \leq d$, and $\mby^s \sim \mbox{Uniform}(-1,1)$ otherwise.
\item Trigonometry $(\mbx,\mby)$: $\mby^s =sin(2\pi\mbx^s)$ for each $s \leq d$, and $\mby^s \sim \mbox{Uniform}(-1,1)$ otherwise.
\end{itemize}
We set $p=50$ and $n=20$ in linear, and $p=30$ and $n=100$ in trigonometry. For each $d=1,\ldots,10$, we repeat the experiment 1000 times, run the test with all methods, and plot the testing power at a type 1 error level of $\alpha=0.05$ in Figure~\ref{fig3}. In these settings, as the number of marginally dependent dimensions $\delta(X,Y)$ increases, all testing methods eventually achieve a testing power of $1$. Among these, the maximum test consistently outperforms the average test, and the original test has the lowest power. 

\section{Real Data}

This experiment aimed to investigate the presence of any dependency between the abundance levels of peptides in human plasma and the occurrence of cancers. Selected Reaction Monitoring (SRM) was employed as a targeted quantitative proteomics technique for measuring protein and peptide abundance in complex biological samples \citep{PMID21248225}. A prior study utilized SRM to identify a total of $318$ peptides from a total of $98$ individuals, among whom $33$ were normal subjects, $10$ had pancreatic cancer, $24$ had colorectal cancer, and $28$ had ovarian cancer \citep{Wang2017}. Consequently, $\Xn$ represents the sample peptide levels with $p=318$. The data is publicly available on a GitHub repository in MATLAB format\footnote{\url{https://github.com/neurodata/MGC-paper/blob/master/Data/Preprocessed/proteomics.mat}}.

We performed independence tests based on various combinations, including utilizing the entire sample dataset, where $\Yn$ represents a label vector indicating the cancer type each subject has. Other test scenarios included: distinguishing normal individuals from others, where $\Yn$ is a label vector with normal subjects as $0$ and all others as $1$; distinguishing pancreatic from colorectal cancer, with $\Yn$ as a label vector where pancreatic subjects are labeled as $0$, colorectal subjects as $1$, and others as unused; and so forth. Note that categorical response variables can be directly incorporated in dependence testing. Binary variables, for instance, can be numerically encoded (e.g., as 0 and 1), while multi-category variables can be handled via one-hot encoding \citep{Zhang2025, DCorMANOVA}.

A previous study \citep{MGC} indicated that all such testing combinations should yield significant p-value. Table~\ref{t1} presents the test statistics and p-values for the maximum distance correlation, average distance correlation, and original distance correlation. While the original distance correlation performed well, there were three combinations where it failed to detect dependence: pancreatic vs others, colorectal vs others, and pancreatic vs colorectal. This may not be surprising, given that there were only $10$ subjects with pancreatic cancer, which is the smallest group in the dataset, and colorectal cancer is the second smallest group. 

The maximum correlation yielded significant results overall, except in three testing combinations involving pancreatic cancer. As the chi-square test for the maximum correlation may be overly conservation for small sample sizes, in these three cases, we further conducted permutation tests using $100$ random permutations (results reported in brackets), and the results improved significantly.

The average correlation yielded significant p-values in almost all combinations, as its chi-square test is robust against small sample sizes. The only exception was the test for pancreatic vs. colorectal cancer, where it did not yield a significant result, while the maximum method is very close to significance. Its underperformance in this case suggests that $\delta(X,Y)$, the number of marginally dependent dimensions, could be very small, leading to its lack of sensitivity.

\begin{table*}[!ht]
\small
\centering
\caption{Results for cancer peptide testing. We consider $11$ different combinations of the sample data to test significant relationship, and use the original distance correlation, maximum distance correlation, and average distance correlation for testing. The statistic and the p-value are reported. }
\label{t1}%
\begin{tabular}{|l|c||c|c||c|c||c|c|}
\hline
Combination & n & $\Dcor$ & p-val & $\MDcor$ & p-val & $\ADcor$ & p-val  \\
\hline
All data & 98 & $0.26$ & $<0.001$&  $0.30$ & $<0.001$ & $0.10$ & $<0.001$   \\
\hline 
Norm vs Others & 98 & $0.13$ & $<0.001$&  $0.23$ & $<0.001$ & $0.055$ & $<0.001$   \\
PANC vs Others & 98 & $0.007$ & $0.19$&  $0.11$ & $0.18 (0.07)$ & $0.003$ & $0.01$   \\
COLO vs Others & 98 & $0.022$ & $0.07$&  $0.15$ & $0.02 $ & $0.013$ & $<0.001$   \\
OVAR vs Others & 98 & $0.35$ & $<0.001$&  $0.47$ & $<0.001$ & $0.15$ & $<0.001$   \\
\hline 
Norm vs. PANC & 43 &$0.080$ & $0.035$&  $0.50$ & $<0.001$ & $0.024$ & $<0.001$   \\
Norm vs. COLO  & 57 &$0.077$ & $0.021$&  $0.36$ & $<0.001$ & $0.029$ & $<0.001$   \\
Norm vs. OVAR & 61 &$0.41$ & $<0.001$&  $0.49$ & $<0.001$ & $0.18$ & $<0.001$   \\
\hline 
PANC vs. COLO & 34 &$-0.030$ & $1$&  $0.31$ & $0.19 (0.06)$ & $-0.01$ & $0.99$   \\
PANC vs. OVAR & 38 &$0.17$ & $0.005$&  $0.25$ & $0.31 (0.02)$ & $0.06$ & $<0.001$   \\
COLO vs. OVAR & 52 &$0.25$ & $<0.001$&  $0.35$ & $<0.001$ & $0.12$ & $<0.001$   \\
\hline 
\end{tabular}
\end{table*}

\section{Conclusion}

In this paper, we propose the maximum and average distance correlations using pairwise, unbiased, and marginal distance correlations. This formulation facilitates the understanding of their consistency properties, relative advantages, limiting distributions, and enables the use of chi-square-based tests. The numerical experiments further confirm our findings and shed light on their practical usages.

Beyond their theoretical appeal, these methods are highly applicable to real-world problems, particularly in biomedical and biological sciences. In such fields, researchers are often interested in detecting complex, nonlinear, and high-dimensional dependencies, such as between gene expression profiles and disease subtypes, or between multi-omics data and clinical outcomes. Our methods are well suited to these tasks, as they operate in a fully nonparametric, model-free framework, require minimal assumptions, and are robust to data heterogeneity.

Moreover, the ability to decompose dependence through marginal statistics (and aggregate them meaningfully) offers practical interpretability and flexibility for applied scientists. Compared to classical approaches rooted in covariance-based inference (e.g., the Wishart distribution under multivariate normality), our approach targets a broader class of multivariate dependence structures.

Taken together, we believe the proposed distance correlation framework offers a valuable statistical tool for both theoretical development and applied data analysis.


\section*{Declarations}

\begin{itemize}
\item Availability of data and materials: The datasets generated and/or analysed during the current study are available in the Github repository, [https://github.com/neurodata/mgc-matlab].
\item Author Contributions: C.S.: conceptualization, method, theory, experiments, writing; Y.D.: conceptualization, theory, discussion. All authors have read and agreed to the published version of the manuscript.
\item Funding: This work was supported in part by the National Science Foundation DMS-1921310 and DMS-2113099, the University of Delaware Data Science Institute Seed Funding Grant, and the Defense Advanced Research Projects Agency's L2M program FA8650-18-2-7834.
\item Acknowledgments: We sincerely thank the editor and anonymous reviewers for providing valuable and timely feedback that significantly improved the paper.
\end{itemize}


\bibliographystyle{ieeetr}
\bibliography{mgc}

 \clearpage
\setcounter{figure}{0}
\setcounter{theorem}{0}
\setcounter{section}{0}
\renewcommand{\thealgorithm}{C\arabic{algorithm}}
\renewcommand{\thetheorem}{A\arabic{theorem}}
\renewcommand{\thedefinition}{B\arabic{definition}}
\renewcommand{\thefigure}{E\arabic{figure}}
\renewcommand{\thesection}{A\arabic{section}}
\renewcommand{\thesubsection}{\thesection.\arabic{subsection}}
\renewcommand{\thesubsubsection}{\thesubsection.\arabic{subsubsection}}
\pagenumbering{arabic}
\renewcommand{\thepage}{\arabic{page}}

\bigskip
\begin{center}
{\large\bf APPENDIX}
\end{center}

\section{Technical Prerequisite}

We first cite the main theorem of \cite{DCorFast}, which is necessary for subsequent proofs. It holds under the assumption of independence between $\mbx$ and $\mby$, finite-moments of $F_{XY}$, an increasing sample size $n$, and regardless of $p,q$ nor the specific distribution of $F_{XY}$.\\

\begin{theorem}
\label{thm0}
For sufficiently large $n$, there exists $\alpha>0$ such that
\begin{align*}
\mc{N}(0,2) \preceq_{\alpha} n \cdot \Dcor(\mathbf{X},\mathbf{Y}) \preceq_{\alpha} \chi_{1}^{2} - 1
\end{align*}
regardless of the metric choice or marginal distributions. \\
\end{theorem}

Here, the notation $\preceq_{\alpha}$ means upper tail dominance in distribution, defined as follows:\\

\begin{definition}
Given two random variables $U$ and $V$, we say $U$ dominates $V$ in upper tail at probability level $\alpha$, or equivalently $V \preceq_{\alpha} U$, if and only if
\begin{align*}
F_{V}(z) \geq F_{U}(z)
\end{align*}
for all $z \geq F_{U}^{-1}(1-\alpha)$, where $F_{V}$ and $F_{U}$ are cumulative distribution functions of random variables $V$ and $U$ respectively.\\
\end{definition}

\section{Theorem Proofs}
\thmOne*
\begin{proof}
When $\delta(X,Y)=0$, $F_{X^s Y^t} = F_{X^s}F_{Y^t}$ for any pair of $(s,t)$, leading to:
\begin{align*}
\Dcor(\Xn^s,\Yn^t) \stackrel{n \rightarrow \infty}{\rightarrow} 0.
\end{align*}
Consequently, when $p$ and $q$ are fixed, both the maximum distance correlation and average distance correlation are asymptotically $0$.

If $pq$ increases together with $n$, there can be an infinite number of marginal correlations. The average distance correlation still converges to $0$ by law of large numbers, as it represents the mean of $pq$ marginal correlations, all of which converge to $0$.

However, the maximum distance correlation may be influenced, and a careful analysis is required to determine whether the convergence still holds in probability. For any $\epsilon > 0$, it suffices to prove
\begin{align*}
&Prob( \MDcor(\Xn,\Yn) < \epsilon) \\
&\geq Prob( n \cdot \Dcor(\Xn^s,\Yn^t) < n \epsilon)^{pq}\\
&\geq F_{\chi_{1}^2-1}^{pq}(n \epsilon) \\
& \rightarrow 1.
\end{align*}
Here, the second line follows from basic order statistics, and the third line follows from the existing dominance results by Theorem~\ref{thm0}.

Next, consider the asymptotic expansion of the standard normal tail probability:
\[
1 - \Phi(x) = \frac{1}{\sqrt{2\pi}x} e^{-x^2/2} \left(1 - \frac{1}{x^2} + \frac{3}{x^4} - \cdots \right), \quad \text{as } x \to \infty,
\]
where $\Phi(x)$ is the standard normal CDF. Applying this to our setting, where
\[
F_{\chi_1^2 - 1}^{pq}(n\epsilon) = \left[ 2\Phi(x) - 1 \right]^{pq},
\]
we immediately have
\begin{align*}
F_{\chi_1^2 - 1}^{pq}(n\epsilon)
= \left(1 - \frac{2}{\sqrt{2\pi}x} e^{-x^2/2} + o\left(\frac{e^{-x^2/2}}{x^3}\right)\right)^{pq} =(1-\delta_n)^{pq},
\end{align*}
where \( x = \sqrt{n\epsilon + 1} = O(\sqrt{n}) \), which increases to infinity at any fixed $\epsilon$. 

We aim to show $(1-\delta_n)^{pq} \rightarrow 1$. It suffices for $pq \delta_n \rightarrow 0$. This is because:
\[
(1 - \delta_n)^{pq} = \exp\left( pq \cdot \log(1 - \delta_n) \right).
\]
As \( \delta_n \to 0 \), we use the Taylor expansion \( \log(1 - \delta_n) = -\delta_n + o(\delta_n) \), so
\[
(1 - \delta_n)^{pq} = \exp\left( -pq \cdot \delta_n + o(pq \cdot \delta_n) \right).
\]
If \( pq \cdot \delta_n \to 0 \), then the exponent tends to zero, and hence
\[
(1 - \delta_n)^{pq} \to 1.
\]

To ensure $pq \delta_n \rightarrow 0$, a sufficient condition would be
\[
pq \cdot \frac{e^{-x^2/2}}{x} \to 0.
\]
To see why, if the above holds, then 
\[
pq \cdot \frac{2}{\sqrt{2\pi}x} e^{-x^2/2} \to 0 
\quad \text{and} \quad 
pq \cdot o\left(\frac{e^{-x^2/2}}{x^3}\right) \to 0,
\]
and hence,
\[
pq \cdot \left( \frac{2}{\sqrt{2\pi}x} e^{-x^2/2} + o\left(\frac{e^{-x^2/2}}{x^3} \right) \right) \to 0.
\]

Finally, to satisfy
\[
pq \cdot \frac{e^{-x^2/2}}{x} \to 0
\]
with \( x = \sqrt{n\epsilon + 1} = O(\sqrt{n}) \), we observe that
\[
\frac{e^{-x^2/2}}{x} = O\left( \frac{e^{-n/2}}{\sqrt{n}} \right).
\]
Therefore, a sufficient condition is
\[
pq = o\left( \sqrt{n} \cdot e^{n} \right).
\]
\end{proof}

\thmTwo*
\begin{proof}
When $\delta(X,Y)> 0$, there exists at least one pair of $(s,t)$ such that 
\begin{align*}
\Dcor(\Xn^s,\Yn^t) \stackrel{n \rightarrow \infty}{\rightarrow} c > 0.
\end{align*}
Therefore, the maximum distance correlation shall converge to a positive constant.

This also holds true for the average distance correlation when $pq$ is fixed. However, when $pq$ also increases to infinity (at any rate relative to $n$), and $\delta(X,Y)=o(pq)$, then:
\begin{align*}
|\ADcor(\Xn,\Yn)|&=|\sum\limits_{s \in [p],t\in[q]}\Dcor(\Xn^s,\Yn^t)|/pq \\
 &\leq \delta(X,Y) / pq \\
 &\rightarrow 0,
\end{align*}
where the second line follows because each marginal correlation is bounded in $[-1,1]$. 
\end{proof}

\thmThree*
\begin{proof}
For each $s \in [p], t \in [q]$, we apply the upper-tail dominance property from Theorem~\ref{thm0} to each marginal distance correlation, i.e.,
\begin{align*}
n \cdot \Dcor(\Xn^s,\Yn^t) &\preceq_{\alpha} \chi_{1}^{2}-1
\end{align*}
for sufficiently large $n$ and sufficiently small $\alpha$. 

By order statistics of independent random variables, we can establish the distribution of the maximum distance correlation as follows:
\begin{align*}
&F_{n \cdot \MDcor(\Xn,\Yn)+1} (z) \\
=& \prod_{s \in [p],t\in[q]}Prob(n \cdot \Dcor(\Xn^s,\Yn^t)+1 \leq z) \\
 \geq& F_{\chi_{1}^{2}}^{pq}(z) 
\end{align*}
Consequently, we have $n \cdot \MDcor(\Xn,\Yn)+1 \preceq_{\alpha} U$.
\end{proof}

\thmFour*
\begin{proof}
First,
\[
n \sqrt{pq} \cdot \ADcor(\Xn, \Yn)
= \frac{1}{\sqrt{pq}} \sum_{s=1}^p \sum_{t=1}^q n \cdot \Dcor(\Xn^s, \Yn^t).
\]

As mentioned in the background section, the limiting null distribution of each marginal distance correlation satisfies
\begin{align*}
n \cdot \Dcor(\Xn^s,\Yn^t) &\stackrel{D}{\rightarrow} \sum\limits_{i,j=1}^{\infty} w_{ij} (\mc{N}_{ij}^{2}-1),
\end{align*}
where the weights satisfy $w_{ij} \in [0,1]$ and $\sum\limits_{i,j=1}^{\infty} w_{ij}^{2} = 1$, and $\mc{N}_{ij}$ are independent standard normal distribution. Therefore, as $n \rightarrow \infty$, each $n \cdot \Dcor(\Xn^s,\Yn^t)$ has an expected value of $0$ and a variance of
\[
\operatorname{Var} \left( \sum_{i,j} w_{ij} ( \mathcal{N}_{ij}^2 - 1 ) \right) = 2 \sum_{i,j} (w_{ij})^2 = 2.
\]

Furthermore, since this limit is a (centered) quadratic form in independent sub-Gaussian random variables, it satisfies uniform exponential tail bounds. Specifically, there exists a constant \( C > 0 \) such that
\[
\sup_{s,t,n} \mathbb{E} \left[ \exp \left( C \cdot \left| n \cdot \Dcor(\Xn^s, \Yn^t) \right| \right) \right] < \infty.
\]

Under the given assumptions, the marginal statistics \( n \cdot \Dcor(\Xn^s, \Yn^t) \) are independent across \( s \in [p] \), \( t \in [q] \). Then we may view the full sum as a triangular array:
\[
Z_n := \frac{1}{\sqrt{pq}} \sum_{s=1}^p \sum_{t=1}^q \left( n \cdot \Dcor(\Xn^s, \Yn^t) \right).
\]
Each summand has mean zero and variance 2, so the total variance of \( Z_n \) is
\[
\operatorname{Var}(Z_n) = \frac{1}{pq} \cdot pq \cdot 2 = 2.
\]

To apply the triangular array Central Limit Theorem (e.g., Lindeberg-Feller CLT), it suffices to verify the Lindeberg condition:
\[
\forall \varepsilon > 0, \quad \lim_{n \to \infty} \frac{1}{pq} \sum_{s,t} \mathbb{E} \left[ \left( n \cdot \Dcor(\Xn^s, \Yn^t) \right)^2 \cdot \mathbf{1} \left\{ \left| n \cdot \Dcor(\Xn^s, \Yn^t) \right| > \varepsilon \sqrt{pq} \right\} \right] = 0.
\]

This condition is satisfied due to the exponential tail bounds established above, which ensure that large deviations decay faster than any polynomial in \( pq \). Therefore, the contribution of rare large terms becomes negligible as \( pq \to \infty \). Hence, the Lindeberg CLT applies, and we conclude:
\[
n \sqrt{pq} \cdot \ADcor(\Xn, \Yn) \xrightarrow{D} \mathcal{N}(0, 2).
\]
\end{proof}

\thmFive*
\begin{proof}
The validity of these tests can be directly deduced from Theorem~\ref{thm3} and Theorem~\ref{thm4}. For the maximum statistic, owing to upper-tail dominance, the p-value obtained through the chi-square-based test is always greater than or equal to the p-value derived from the true null. Consequently, the test is valid and tends to be more conservative than the permutation test. In the case of the average statistic, the p-value produced by the chi-square-based test converges to the p-value generated by the true null. Hence, the test is valid and matches the p-value of the permutation test for sufficiently large $n$, $p$, and $q$.
\end{proof}

\section{Standard Permutation Test}

While the main paper uses the chi-square test, it can be slightly conservative when the sample size is small. An alternative is the standard permutation test, although more computationally expensive, tends to be more accurate and is applicable to both the average and maximum distance correlation statistics. Taking $\MDcor(\mathbf{X}, \mathbf{Y})$ as an example:

Given paired data $\mathbf{X}$ and $\mathbf{Y}$, we generate $R$ independent permutations of $\mathbf{Y}$ by randomly shuffling the sample indices. Denote the $r$-th permuted data as $\pi_r(\mathbf{Y})$, and compute the test statistic $\MDcor(\mathbf{X}, \pi_r(\mathbf{Y}))$ for each $r = 1, \dots, R$.

The permutation-based p-value is then given by:
\begin{align*}
\mbox{pval}= \frac{1}{R}\sum_{r=1}^{R} 1\{\MDcor(\mathbf{X},\pi_r(\mathbf{Y}))> \MDcor(\mathbf{X},\mathbf{Y})\}.
\end{align*}

\section{Additional Simulation}
\label{appenSim}

Here, we present a special example where the random vectors $X$ and $Y$ are dependent, yet each individual dimension of $X$ is independent of $Y$. As a result, neither the average nor the maximum marginal distance correlation exhibits meaningful testing power. In contrast, the original distance correlation remains consistent and achieves perfect power as the sample size increases.

We use an XOR-like construction with categorical variables: let $X^1 \sim \text{Bernoulli}(0.5)$ and $Y \sim \text{Bernoulli}(0.5)$ be independent, and define $X^2 = X^1 \oplus Y$, where $\oplus$ denotes the exclusive OR operation. Specifically, $X^2 = 1$ if $X^1 \neq Y$, and $0$ otherwise. By construction, $X^1$ is independent of $Y$, and one can verify that $X^2$ is also independent of $Y$ by enumerating all possible combinations. However, the pair $X = [X^1, X^2]$ determines $Y$ uniquely (since $Y = X^1 \oplus X^2$), implying that $X$ and $Y$ are dependent, despite each marginal of $X$ being independent of $Y$.

In this experiment, we fix the sample size $n = 100$ and compute the original distance correlation, maximum distance correlation, and average distance correlation, each using their respective chi-squared test. Over 1000 Monte Carlo replicates, the empirical power at a type I error level of $0.05$ is: 1.000, 0.046, and 0.048, respectively.

These results remain unchanged even when increasing the sample size $n$, or when replacing distance correlation with HSIC. Therefore, this example demonstrates that there exists dependence structures where any marginal-based dependence measure fails, while the original distance correlation or HSIC remains consistent and achieves perfect testing power.

\end{document}